\documentclass[conference]{IEEEtran}
\IEEEoverridecommandlockouts

\usepackage{cite}
\usepackage{amsmath,amssymb,amsfonts}
\usepackage{algorithmic}
\usepackage{graphicx}
\usepackage{textcomp}
\usepackage{xcolor}
\usepackage{url}
\usepackage{mathptmx} 
\usepackage{newtxmath}
\usepackage{times}

\usepackage{multirow}
\usepackage{here}
\usepackage{longtable}
\usepackage{float}
\usepackage{multirow}
\usepackage{comment}
\usepackage{siunitx}
\usepackage {diagbox}
\usepackage{fancybox}
\usepackage{ascmac}

\def\BibTeX{{\rm B\kern-.05em{\sc i\kern-.025em b}\kern-.08em
    T\kern-.1667em\lower.7ex\hbox{E}\kern-.125emX}}

\begin{document}

\title{Training Dialogue Systems by AI Feedback \\for Improving Overall Dialogue Impression}

\author{
\IEEEauthorblockN{Kai Yoshida$^{1,2}$, Masahiro Mizukami$^{3}$, Seiya Kawano$^{2,1}$,Canasai Kruengkrai$^{2}$,}
\IEEEauthorblockN{Hiroaki Sugiyama$^{3}$,Koichiro Yoshino$^{4,2,1}$}
\IEEEauthorblockA{yoshida.kai.yf1@is.naist.jp, masahiro.mizukami@ntt.com, seiya.kawano@riken.jp,}
\IEEEauthorblockA{canasai.kruengkrai@riken.jp, h.sugi@ieee.org, yoshino.k.ai@m.titech.ac.jp}
\IEEEauthorblockA{
$^{1}$Nara Institute of Science and Technology, Japan
$^{2}$Guardian Robot Project, RIKEN, Japan\\
$^{3}$NTT Communication Science Laboratories, Japan
$^{4}$Institute of Science Tokyo, Japan
}
}

\maketitle

\begin{abstract}
To improve user engagement during conversations with dialogue systems, we must improve individual dialogue responses and dialogue impressions such as consistency, personality, and empathy throughout the entire dialogue.
While such dialogue systems have been developing rapidly with the help of large language models (LLMs), reinforcement learning from AI feedback (RLAIF) has attracted attention to align LLM-based dialogue models for such dialogue impressions.
In RLAIF, a reward model based on another LLM is used to create a training signal for an LLM-based dialogue model using zero-shot/few-shot prompting techniques.
However, evaluating an entire dialogue only by prompting LLMs is challenging.
In this study, the supervised fine-tuning (SFT) of LLMs prepared reward models corresponding to 12 metrics related to the impression of the entire dialogue for evaluating dialogue responses.
We tuned our dialogue models using the reward model signals as feedback to improve the impression of the system.
The results of automatic and human evaluations showed that tuning the dialogue model using our reward model corresponding to dialogue impression improved the evaluation of individual metrics and the naturalness of the dialogue response.
\end{abstract}
\renewcommand{\thefootnote}{\fnsymbol{footnote}}
\renewcommand{\thefootnote}{\arabic{footnote}}


\section{Introduction}
Recent advancements in large language models (LLMs) have achieved models that respond naturally to given input sentences~\cite{lee-etal-2023-prompted}.
Current LLMs such as ChatGPT~\cite{openai-chatgpt,gpt4-2023-report} and Bard~\cite{Manyika-bard-2023} has a framework to adapt their output as people like using human feedback, called reinforcement learning from human feedback (RLHF)~\cite{long-rlhf-2022,daniel-rlhf-2020,bai-2022-helpful,Christiano-2017-deep}.
In RLHF, human evaluators give binary or ranking feedback to the LLM's output, which is used to update the LLMs word generation process via reinforcement learning.
High-quality RLHF assumes a variety of feedback, which requires either a correspondingly large number of evaluators or a large, high-quality human evaluation dataset~\cite{lee-rlaif-2023}.
For this reason, a recent approach called reinforcement learning from AI feedback (RLAIF)~\cite{lee-rlaif-2023,cheng-aif-2024,Bai-constitutional-2022,minae-reward-2023,singh-scaling-2023,pmlr-v202-hu23e} has been developed in which the role of reward feedback, which humans should perform, is assigned to a reward model typically based on LLMs.
In other words, we prepare an LLM that is different from the dialogue model for the feedback.

Most existing studies of RLAIF assume a one-to-one evaluation given to the generated text. 
For example, Cheng et al.~\cite{cheng-aif-2024} used model training based on feedback from LLMs for inference and translation, and Lee et al.~\cite{lee-rlaif-2023} and Bai et al.~\cite{Bai-constitutional-2022} aimed to remove sentences that are harmful to the user.

When trying to provide feedback to a dialogue model using an LLM or reward model, it is important to evaluate not only each response but also the user's overall dialogue impression.
For example, improving the dialogue system's consistency of responses, personality, and empathy will improve the users' dialogue experiences.
The reward model must evaluate the overall dialogue impression using contexts up to that point. 
The evaluation cost becomes even higher when considering the consistency of the overall dialogue context and impression evaluation.
Several existing studies of LLMs tackled the problem of evaluating overall dialogue impressions, including FED~\cite{mehri-eskenazi-2020-unsupervised}, MEEP~\cite{ferron-etal-2023-meep}, INCAHARACTER~\cite{wang-etal-2024-incharacter}, LLM-Eval~\cite{lin-chen-2023-llm}, G-Eval~\cite{liu-etal-2023-g}, LLM-as-a-Judge~\cite{zheng-judge-2023}. 
However, it is difficult to give an appropriate evaluation from a specific evaluation viewpoint for the whole dialogue only by zero-shots and few-shots prompting of LLMs, especially for using them on tuning dialogue models to each evaluation viewpoint.

This study investigates the best strategy to tune dialogue models to evaluate viewpoints for overall dialogue impression scores. We compare reward models based on prompting and supervised fine-tuning using a small dialogue dataset annotated with overall dialogue impression scores. 
Specifically, using the prepared dialogue evaluation data, a regression task is trained and used to answer 12 different overall dialogue impressions. 
We tune the dialogue models to improve the output from these reward models. 
Our experimental results showed that tuning dialogue models using such reward models for overall dialogue impressions achieved the best performance on both automatic and human evaluations.

\section{Reward Model for Dialogue Impression}\label{sec:reward}
In this study, we prepared and used a reward model to evaluate the LLM-based dialogue model.
It isn't easy to evaluate overall dialogue impressions by zero-shot/few-shot prompting; therefore, in this study, we apply SFT to a reward model based on LLMs. 
In this section, we explain this training method of the reward model and the difference with the general prompting method in correlation with human scores. 

\subsection{Dialogue Evaluation Task and Data}
Evaluation metrics related to the overall dialogue impression include, for example, consistency and empathy. 
This study will use the 12 metrics annotated on JTransformer-Eval~\cite{sugiyama-jtrans-2022} to evaluate the overall dialogue impression. 
JTransformer-Eval is a dataset containing dialogues between the japanese-dialog-transformer~\cite{sugiyama-jtrans-2022} (system) and users (humans) and their evaluation labels.
The 12 evaluation metrics are shown in Table~\ref{tab:difinition_aif}.
Each is rated on an 11-point scale from 0 to 10 by human annotators.
\begin{table}[t]
\footnotesize
    \centering
    \caption{Evaluation metrics and corresponding questionnaires}
    \begin{tabular}{r|p{6.5cm}}
    Metric name & Questionnaire \\ \hline
Agency & I felt that the system was speaking from its perspective \\ 
Attentiveness & The system was interested in me and was actively trying to talk with me \\ 
Consistency & The system's utterances were consistent and coherent \\ 
Ease & Continuing the dialogue was easy \\
Empathetic & I was able to empathize with the system's utterances \\ 
Emotion & I felt that the system had feelings \\
Enjoyability & I enjoyed interacting with the system \\ 
Humanness & The system's utterances were humanlike and natural \\  
Personality & I could sense the system’s personality and character \\ 
Respeak & I want to talk with this system again \\ 
Topic & I felt that the system had a topic it wanted to discuss \\ 
Trust & I felt that what the system said was trustworthy \\ 
    \end{tabular}
    \label{tab:difinition_aif}
\end{table}
All dialogues consist of 32 turns, all set up so that the system initiates them and prompts the end of the dialogues at a certain turn. 
The dataset contains 1,600 dialogue sessions.

\subsection{Reward Model Training}\label{sec:rm_train}
The reward model evaluates the overall dialogue impression on a scale of 0-10, with the dialogue context $C_i$, its response $R_i$, and the evaluation score $S_{i, E_j}$ corresponding to a metric $E_j$ as inputs. 
Here, $i$ represents the index identifying each dialogue sample, and $j$ represents the index identifying each evaluation metric.
We fine-tune an LLM to regress the score.
A linear layer is added to the final layer of the original model to train this model as a regression model rather than a generative model.
We also train a single model to evaluate 12 different metrics.
In training, the mean squared error was used as the loss function; the epoch was set to 10, and the batch size was 16 for eight devices.
Therefore, the final batch size is 128. 
We tuned llm-jp/llm-jp-3-1.8b-instruct\footnote{\url{https://huggingface.co/llm-jp/llm-jp-3-1.8b-instruct}} and tokyotech-llm/Swallow-7b-instruct-hf\footnote{\url{https://huggingface.co/tokyotech-llm/Swallow-7b-instruct-hf}}.
The final model was selected by validation loss.

\subsection{Evaluation of Reward Models}
To investigate the effect of our model tuning, we compared four different reward models: an instruction-tuned model with prompting (w/o SFT), GPT-3.5-turbo with prompting (GPT-3.5), and a tuned model (w/ SFT 1.8b and 7b) described in Section~\ref{sec:rm_train} using JTransfomer-Eval.
We used the average probability of generating score labels for the evaluation prompt multiplied by each score label.
We used a test set of JTransofmer-Eval data divided into train, dev, and test as 8:1:1.

The Spearman correlations between the human scores and predicted scores in the test set are shown in Table~\ref{tab:aif_acc}.
Correlations that could not be computed because only one type of score was output during inference are replaced by “-.”
\begin{table}[t]
\small
\centering
\caption{Correlation between evaluation model and human evaluation}
\begin{tabular}{cSSSS}
\hline
Metrics & w/o SFT & {GPT-3.5} & {w/ SFT 1.8b} & {w/ SFT 7b} \\ \hline
Agency & 0.0 & 0.21 & 0.44 & \textbf{0.79} \\ \hline
Attentiveness & 0.11 & 0.30 & 0.41 & \textbf{0.85} \\ \hline
Consistency & 0.14 & 0.40 & 0.53 & \textbf{0.84} \\ \hline
Ease & 0.07 & 0.29 & 0.36 & \textbf{0.79}\\ \hline
Empathetic & 0.10 & 0.41 & 0.46 & \textbf{0.85} \\ \hline
Emotion & 0.08 & 0.26 & 0.36 & \textbf{0.82} \\ \hline
Enjoyability  & 0.03 & 0.24 & 0.42 & \textbf{0.79} \\ \hline
Humanness & 0.08 & 0.32 & 0.32 & \textbf{0.77} \\ \hline
Personality & -0.06 & 0.27 & 0.40 & \textbf{0.76} \\ \hline
Respeak & {-} & 0.25 & 0.44 & \textbf{0.85} \\ \hline
Topic & -0.15 & 0.33 & 0.42  & \textbf{0.77} \\ \hline
Trust & {-} & 0.18 & 0.55 & \textbf{0.89} \\ \hline
\end{tabular}
\label{tab:aif_acc}
\end{table}
The model without SFT tended to output the same score for different samples, confirming a weak correlation.
GPT-3.5 achieves weak correlations of around 0.2 to 0.4.
Our 1.8b SFT model achieved positive correlations of around 0.3 to 0.5. 
Furthermore, the 7b SFT model achieved strong positive correlations of around 0.8 for all evaluation metrics.
In particular, Trust achieved a very strong correlation close to 0.9.

This experiment confirmed that it is difficult to evaluate the overall dialogue impression with an LLM that simply performs prompting and that the SFT of the model corresponding to each evaluation metric is necessary. 
The model trained by 7b SFT is used in the following sections as the reward model for tuning the dialogue model.

\section{Dialogue Optimization with AI Feedback}
This section provides an overview of our system implementation and its evaluation methods.
In RLAIF, 24 combinations of 12 evaluation values and two training methods are tested, and the training results are compared using automatic and human evaluation.

\subsection{System Overview}
\begin{figure}[t]
\centering
    \includegraphics[width=\linewidth]{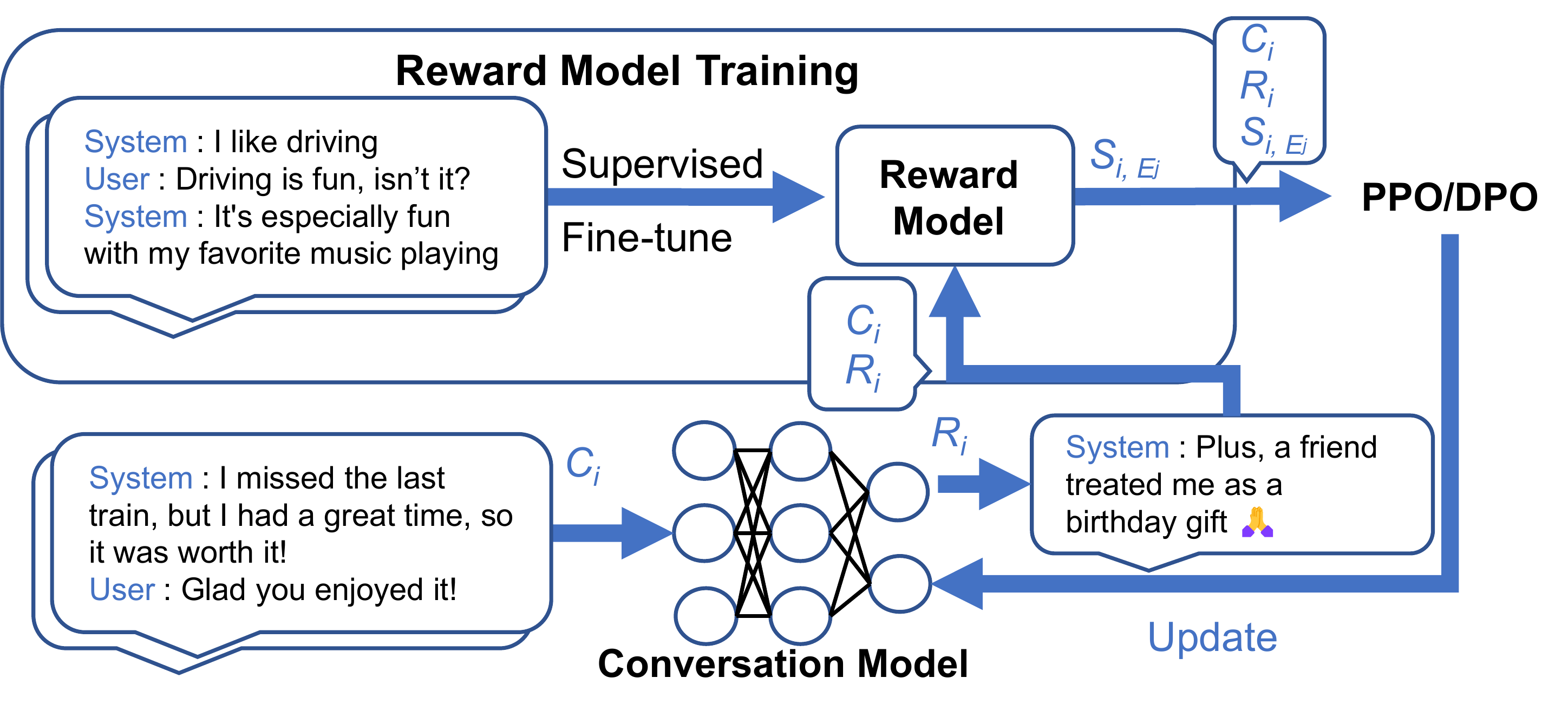}
    \caption{Training of reward model and implementation of PPO and DPO, where $C_i$ is dialogue context, $R_i$ is generated response, the evaluation score $S_{i, E_j}$ corresponding to a metric $E_j$ as inputs}
    \label{fig:rlaif}
\end{figure}

An overview of the training to be conducted in this study is shown in Fig.~\ref{fig:rlaif}.
In training, a score $S_{i, E_j}$ is assigned to the dialogue context $C_i$ and LLMs response $R_i$ to $C_i$ using a reward model corresponding to a specific dialogue impression metrics $E_j$.
The model is then updated based on $(C_i, R_i, S_{i, E_j})$.

\subsection{LLM Training from AI Feedback}
For training, LLMs based on the output of the reward model include proximal policy optimization (PPO)~\cite{schulman-ppo-2017} and direct policy optimization (DPO)~\cite{rafailov-dpo-2024}.
PPO updates the model by evaluating the reward model against the generation of LLMs during the training loop. In contrast, DPO uses a pre-processed dataset and does not provide feedback to the generation process using the reward model during the training loop.
In DPO pre-processing, the model generates two types of responses to update the dialogue history, and the reward model evaluates these two sentences for the dialogue history.
The generated sentence with the higher evaluation value is then used for training as accepted and the lower as rejected.

To check the generalization performance of RLAIF, two models were used in training: cyberagent/calm2-7b-chat\footnote{\url{https://huggingface.co/cyberagent/calm2-7b-chat}} (calm) and rinna/youri-7b-chat\footnote{\url{https://huggingface.co/rinna/youri-7b-chat}} (youri).
We then trained on 48 combinations of two reinforcement learning methods, two different models, and 12 different evaluation values.
The training was performed using eight A100 80GB GPU devices with a batch size of 32 per device. 
The epoch for PPO was 2, the same as the original RLHF~\cite{long-rlhf-2022}, and 120 was used for DPO.
The final model for DPO was the one with the lowest training loss.

\section{Task Settings}
\subsection{Description of Tasks and Datasets}
This section explains the JEmpathetic Dialogue dataset used for training.
JEmpathetic Dialogue~\cite{sugiyama-jtrans-2022} is a data set of 20,000 four-turn dialogues between systems and humans.
The 32 emotion words expressing certain emotions contained in the original EmpatheticDialogues~\cite{rashkin-etal-2019-towards} were translated into Japanese, and Japanese speakers used them to construct the situation and dialogue sentences.
In this data, 20,000 dialogues and 80,000 utterance pairs were collected.
To implement reinforcement learning, the data was split 8:1:1 for training, validation, and testing.

\begin{table*}[t]
\tiny
\centering
\caption{Results of the automatic evaluation, where AIF is the eleven grades}
\begin{tabular}{c|SS|SS|SS|SS|SS|SS}
 & \multicolumn{6}{c}{calm} & \multicolumn{6}{c}{youri} \\
 & \multicolumn{2}{c}{w/o tuning} & \multicolumn{2}{c}{PPO} & \multicolumn{2}{c}{DPO} &  \multicolumn{2}{|c}{w/o tuning} & \multicolumn{2}{c}{PPO} & \multicolumn{2}{c}{DPO} \\ \cline{2-13}
 & {AIF} & {PPL} & {AIF} & {PPL} & {AIF} & {PPL} & {AIF} & {PPL} & {AIF} & {PPL} & {AIF} & {PPL} \\ \hline
Agency & 5.74 & 42.12 & 5.74 & 42.83 & \textbf{6.50} & \textbf{36.74} & 5.43 & 25.56 & 5.29 & 28.40 & \textbf{5.91} & \textbf{23.36}  \\ \hline
Attentiveness & 5.03 & 39.72 & 5.03 & 41.32 & \textbf{5.66} & \textbf{33.98} & 4.85 & 23.57 & 4.76 & 46.75 & \textbf{5.17} & \textbf{23.43}  \\ \hline
Consistency & 4.38 & 39.01 & 4.38 & 41.62 & \textbf{4.93} & \textbf{31.94} & 4.01 & 22.65 & 4.04 & 63.92 & \textbf{4.46} & \textbf{22.29}  \\ \hline
Ease & 6.71 & 38.74 & 6.70 & 41.46 & \textbf{7.24} & \textbf{34.21} & 6.60 & \textbf{22.28} & 6.22 & 913.52 & \textbf{6.96} & 22.94  \\ \hline
Empathetic & 4.77 & 38.62 & 4.80 & 41.48 & \textbf{5.17} & \textbf{30.41} & 4.82 & 22.14 & 4.85 & 40.98 & \textbf{5.13} & \textbf{19.22}  \\ \hline
Emotion & 4.62 & 38.56 & 4.62 & 40.60 & \textbf{5.32} & \textbf{31.67} & 4.44 & 22.06 & 4.54 & 50.25 & \textbf{5.02} & \textbf{16.77}  \\ \hline
Enjoyability & 5.32 & 38.55 & 5.33 & 41.40 & \textbf{5.93} & \textbf{29.73} & 5.21 & \textbf{21.98} & 5.18 & 20.96 & \textbf{5.54} & 23.92  \\ \hline
Humanness & 6.31 & 38.55 & 6.34 & 42.44 & \textbf{6.80} & \textbf{33.34} & 6.18 & 21.96 & 6.10 & 41.93 & \textbf{6.48} & \textbf{20.85}  \\ \hline
Personality & 6.00 & 38.55 & 6.02 & 41.10 & \textbf{6.59} & \textbf{31.93} & 5.83 & 21.95 & 5.82 & 35.61 & \textbf{6.37} & \textbf{12.53}  \\ \hline
Respeak & 4.81 & 38.55 & 4.85 & 40.54 & \textbf{5.46} & \textbf{32.51} & 4.68 & 21.95 & 4.65 & 23.97 & \textbf{5.10} & \textbf{20.59}  \\ \hline
Topic & 4.92 & 38.55 & 4.92 & 42.14 & \textbf{5.53} & \textbf{34.44} & 4.75 & \textbf{21.95} & 4.93 & 37.61 & \textbf{5.23} & 26.91  \\ \hline
Trust & 4.29 & 38.55 & 4.29 & 41.85 & \textbf{4.76} & \textbf{31.32} & 4.07 & 21.95 & 4.07 & 32.60 & \textbf{4.47} & \textbf{19.04}  \\ \hline
\end{tabular}
\label{tab:automatic_eval}
\end{table*}

\begin{table*}[t]
\tiny
\centering
\caption{Results of the ranking evaluation, where Rank is the rank on a three-point scale and Win is the percentage that were at or above the same order compared to before the study}
\begin{tabular}{c|S|SS|SS|S|SS|SS}
& \multicolumn{5}{c}{calm} & \multicolumn{5}{c}{youri} \\
& \multicolumn{1}{c}{w/o tuning} & \multicolumn{2}{c}{PPO} & \multicolumn{2}{c|}{DPO}  & \multicolumn{1}{c}{w/o tuning} & \multicolumn{2}{c}{PPO} & \multicolumn{2}{c}{DPO} \\ \cline{2-11}
  & Rank & Rank & Win  & Rank & Win & Rank  & Rank & Win & Rank & Win \\ \hline
Agency & 1.76 & \textbf{1.75} & \textbf{0.59} & 1.84 & 0.57 & 1.62 & \textbf{1.58} & \textbf{0.69} & 1.59 & 0.68 \\ \hline
Attentiveness & 1.95 & 1.85 & 0.65 & \textbf{1.55} & \textbf{0.71} & 2.01 & 1.84 & 0.64 & \textbf{1.5} & \textbf{0.78} \\ \hline
Consistency & 1.91 & 1.85 & 0.61 & \textbf{1.69} & \textbf{0.67} & \textbf{1.48} & 1.66 & 0.65 & 1.5 & \textbf{0.68} \\ \hline
Ease & 1.74 & \textbf{1.68} & \textbf{0.58} & 2.01 & 0.47  & \textbf{1.73} & 2.0 & 0.45 & 1.9 & \textbf{0.52} \\ \hline
Empathetic & 2.19 & 1.84 & 0.69 & \textbf{1.44} & \textbf{0.84} & 1.97 & 1.8 & 0.65 & \textbf{1.63} & \textbf{0.71} \\ \hline
Emotion & 2.07 & 1.82 & 0.66 & \textbf{1.47} & \textbf{0.78} & 1.8 & 1.79 & 0.64 & \textbf{1.68} & \textbf{0.65} \\ \hline
Enjoyability & 2.33 & 1.82 & 0.74 & \textbf{1.57} & \textbf{0.79} & 2.0 & \textbf{1.76} & \textbf{0.68} & \textbf{1.76} & 0.6 \\ \hline
Humanness & 1.99 & \textbf{1.73} & \textbf{0.61} & 1.91 & 0.58 & 1.7 & 2.26 & 0.39 & \textbf{1.49} & \textbf{0.64} \\ \hline
Personality & 2.14 & 1.82 & 0.73 & \textbf{1.36} & \textbf{0.83} & 2.01 & 1.71 & 0.68 & \textbf{1.58} & \textbf{0.71} \\ \hline
Respeak & 2.15 & \textbf{1.79} & 0.65 & 1.8 & \textbf{0.66} & 1.87 & 1.82 & 0.63 & \textbf{1.51} & \textbf{0.76} \\ \hline
Topic & 1.97 & \textbf{1.82} & \textbf{0.61} & 1.86 & 0.59 & 1.77 & 1.85 & 0.6 & \textbf{1.36} & \textbf{0.79} \\ \hline
Trust & 2.1 & 1.87 & 0.66 & \textbf{1.56} & \textbf{0.76} & 1.9 & 1.79 & 0.64 & \textbf{1.54} & \textbf{0.74} \\ \hline
\end{tabular}
\label{tab:human_eval_rank}
\end{table*}

\begin{table}[t]
\tiny
\centering
\caption{Results of manual naturalness rating (5 grades)}
\begin{tabular}{c|SSS|SSS}
& \multicolumn{3}{c}{calm} & \multicolumn{3}{c}{youri} \\
  & \multicolumn{1}{c}{w/o tuning} & \multicolumn{1}{c}{PPO} & \multicolumn{1}{c|}{DPO}  & \multicolumn{1}{c}{w/o tuning} & \multicolumn{1}{c}{PPO} & \multicolumn{1}{c}{DPO} \\ \hline
Agency & 2.48 & 2.49 & \textbf{2.85} & 2.12 & \textbf{2.25} & 1.96 \\ \hline
Attentiveness & 2.45 & \textbf{2.66} & 2.64 & 2.33 & 1.67 & \textbf{2.44} \\ \hline
Consistency & 2.42 & \textbf{2.68} & \textbf{2.6} & 1.91 & 1.52 & \textbf{1.92} \\ \hline
Ease & 2.48 & 2.37 & \textbf{2.61} & 1.97 & 1.77 & 1.68 \\ \hline
Empathetic & 2.39 & \textbf{2.75} & \textbf{3.14} & 1.93 & 1.47 & \textbf{1.97} \\ \hline
Emotion & 2.46 & 2.51 & \textbf{2.92} & \textbf{2.03} & 1.68 & 1.99 \\ \hline
Enjoyability & 2.48 & \textbf{2.63} & \textbf{2.63} & 1.95 & 1.71 & 1.89 \\ \hline
Humanness & 2.43 & 2.36 & \textbf{2.6} & \textbf{2.02} & 1.8 & 1.98 \\ \hline
Personality & 2.48 & \textbf{2.71} & 2.51 & \textbf{2.08} & 1.7 & \textbf{2.09} \\ \hline
Respeak & 2.53 & 2.58 & \textbf{2.86} & 2.26 & \textbf{2.57} & 2.36 \\ \hline
Topic & \textbf{2.5} & 2.42 & 2.44 & \textbf{2.13} & 1.54 & 1.77 \\ \hline
Trust & 2.49 & \textbf{2.78} & \textbf{3.08} & 2.17 & 1.91 & \textbf{2.46} \\ \hline
\end{tabular}
\label{tab:human_eval_nat}
\end{table}

\subsection{Evaluation Metrics}
\subsubsection{Automatic Evaluation Method}
In the automatic evaluation, our system evaluates whether LLMs generated sentences have become more fluent compared to the pre-trained model and whether the evaluation values of the dialogue impression to be adapted are reflected, respectively.
Perplexity (PPL) is used to evaluate fluency, and a reward model confirms that the evaluated values reflect the actual AI Feedback (AIF) values.
However, the base model of the post-enhanced training model is used for PPL evaluation, and the PPL of the dialogue context is excluded from the calculation.
In other words, the input dialogue context $C=(c_{0},c_{1},. ,c_{t})$ and the generated sentences $R=(r_{0},r_{1},... ,r_{n})$ are calculated as in Equation ~(\ref{eq:ppl}).
\begin{equation}\label{eq:ppl}
\text{PPL}(R) = \exp\left\{-\frac{1}{n} \sum_{i=0}^{n} \log p_{\theta}(r_{i} \mid C, r_{<i})\right\}
\end{equation}

\subsubsection{Human Evaluation}
In the human evaluation, two aspects are evaluated: whether the response was natural for the dialogue history and whether it reflects the dialogue impression of the adaptation target.
Naturalness is evaluated in five levels (5 being the best and 1 being the worst), and the degree of reflection of the evaluated value is ranked (1 being the best and 3 being the worst) by presenting three types of generated sentences: pre-training, PPO, and DPO, respectively, for each evaluated value. 
However, the rankings could be the same.
The instructions presented to the evaluators were as follows:
\begin{itembox}[l]{Evaluation of naturalness}
For each of the three types of responses, rate on a scale of 1 to 5 whether the response was a natural response to the conversation history. 5 Very natural, 4 Natural, 3 Somewhat undesirable, 2 Unnatural, 1 Very unnatural
\end{itembox}
\begin{itembox}[l]{\parbox{0.8\hsize}{Assessment of the degree of reflection\newline of the evaluation value of dialogue impressions}}
Please rank the questions for each Table~\ref{tab:difinition_aif} value starting with the one you felt the most strongly. 
Those that were felt to the same degree may be in the same order. Please do not consider naturalness or fluency of speech in this evaluation; this evaluation is only based on the criteria for each reward.
\end{itembox}
According to the instructions, one rater evaluated 100 pieces of evaluation data for each metric.

\section{Results}
\subsection{Automatic Evaluation}
Automatic evaluation metrics are used to check whether the training method has increased the rewarded evaluation values and whether the fluency of the generated sentences has been impaired.
The results of the automatic evaluation are shown in Table~\ref{tab:automatic_eval}.
For calm and youri, DPO showed the best results for both AIF and PPL.
In PPO, AIF increased or decreased slightly compared to w/o tuning.
On the other hand, the AIF of DPOs successfully improved by an average of around 0.5.
Regarding PPL, calm achieved the best values for DPOs in all rewards, while youri achieved better values before training in several metrics.

\subsection{Human Evaluation}
The results of the evaluation of manual dialogue impressions are presented in Table~\ref{tab:human_eval_rank}, with naturalness presented in Table~\ref{tab:human_eval_nat}.
Regarding naturalness, DPO is the best in calm, while in youri, pre-training and DPO show superior results.
The improved naturalness of both PPO and DPO compared to pre-training suggests that our training improves the naturalness of responses and reflects evaluation values.
For Rank and Win, DPO has the best results for calm and youri.
The results of automatic and human evaluation suggest that DPO best reflects the naturalness of responses and evaluation values.

\section{Discussion}
For both PPO and DPO, the loss function is different from the cross-entropy during pre-training; DPO improves the PPL as it adapts the model to the base model's generating text, whereas PPO worsens the PPL from the perspective of adapting to the reward model. 
PPO also worsened PPL and did not increase AIF values much compared to w/o tuning, while the human evaluation, their scores for naturalness, and Winning Rate outperformed them.
This suggests that training of response spontaneity and rating values is possible, albeit inferior to DPO.

The evaluation by AIF confirmed that even responses that were natural as a response but did not reflect well on evaluation values of overall dialogue impression would receive a high score if they were natural to some extent.
These were often called dull responses, such as ‘Yes’ or ‘I think it's good.’
This is probably why the PPO's tuning led to such dull response generation.
The scoring of JTransformer-Eval implicitly includes an evaluation of naturalness; 
It is intended to annotate the overall dialogue impression from each perspective.
To better adaptation of the dialogue model, it is important to consider a pure method of evaluating the value of the rewarded evaluation and the diversity of the generated text.

\section{Conclusion}
In this study, we fine-tuned an LLM-based dialogue model to improve the overall dialogue impression by AI feedback. 
We applied SFT to prepare the reward model used to train the dialogue model. We verified whether we can adapt dialogue models for these metrics corresponding to the overall dialogue impression.
For automatic and human evaluation, trained by DPO showed the best results, confirming that our training improves the naturalness of the generated sentences and the evaluation metrics of overall dialogue impressions used as rewards.
On the other hand, some issues were identified, such as the fact that the evaluation model gives high ratings to natural responses even if they are inherently undesirable, such as a dull response, and countermeasures were found to be needed to address these issues.

\section{Acknowledgement}
The content of Section~\ref{sec:reward} is based on joint research conducted by NTT and RIKEN.
\newpage
\bibliographystyle{unsrt}
\bibliography{anthology,j_yourrefs} 

\begin{thebibliography}{10}

\bibitem{lee-etal-2023-prompted}
Gibbeum Lee, Volker Hartmann, Jongho Park, Dimitris Papailiopoulos, and Kangwook Lee.
\newblock Prompted {LLM}s as chatbot modules for long open-domain conversation.
\newblock In Anna Rogers, Jordan Boyd-Graber, and Naoaki Okazaki, editors, {\em Findings of the Association for Computational Linguistics: ACL 2023}, pages 4536--4554, Toronto, Canada, July 2023. Association for Computational Linguistics.

\bibitem{openai-chatgpt}
OpenAI.
\newblock Introducing chatgpt.
\newblock \url{https://openai.com/index/chatgpt}.

\bibitem{gpt4-2023-report}
OpenAI, Josh Achiam, and Steven~Adler et~al.
\newblock Gpt-4 technical report, 2023.

\bibitem{Manyika-bard-2023}
James Manyika.
\newblock An overview of bard: an early experiment with generative ai, 2023.
\newblock \url{https://ai.google/static/documents/google-about-bard.pdf}.

\bibitem{long-rlhf-2022}
Long Ouyang, Jeff Wu, Xu~Jiang, Diogo Almeida, Carroll~L. Wainwright, Pamela Mishkin, Chong Zhang, Sandhini Agarwal, Katarina Slama, Alex Ray, John Schulman, Jacob Hilton, Fraser Kelton, Luke Miller, Maddie Simens, Amanda Askell, Peter Welinder, Paul Christiano, Jan Leike, and Ryan Lowe.
\newblock Training language models to follow instructions with human feedback, 2022.
\newblock arXiv:2203.02155.

\bibitem{daniel-rlhf-2020}
Daniel~M. Ziegler, Nisan Stiennon, Jeffrey Wu, Tom~B. Brown, Alec Radford, Dario Amodei, Paul Christiano, and Geoffrey Irving.
\newblock Fine-tuning language models from human preferences, 2019.
\newblock arXiv:1909.08593.

\bibitem{bai-2022-helpful}
Yuntao Bai, Andy Jones, Kamal Ndousse, Amanda Askell, Anna Chen, Nova DasSarma, Dawn Drain, Stanislav Fort, Deep Ganguli, Tom Henighan, Nicholas Joseph, Saurav Kadavath, Jackson Kernion, Tom Conerly, Sheer El-Showk, Nelson Elhage, Zac Hatfield-Dodds, Danny Hernandez, Tristan Hume, Scott Johnston, Shauna Kravec, Liane Lovitt, Neel Nanda, Catherine Olsson, Dario Amodei, Tom Brown, Jack Clark, Sam McCandlish, Chris Olah, Ben Mann, and Jared Kaplan.
\newblock Training a helpful and harmless assistant with reinforcement learning from human feedback, 2022.
\newblock arXiv:2204.05862.

\bibitem{Christiano-2017-deep}
Paul Christiano, Jan Leike, Tom~B. Brown, Miljan Martic, Shane Legg, and Dario Amodei.
\newblock Deep reinforcement learning from human preferences, 2017.
\newblock arXiv:1706.03741.

\bibitem{lee-rlaif-2023}
Harrison Lee, Samrat Phatale, Hassan Mansoor, Thomas Mesnard, Johan Ferret, Kellie Lu, Colton Bishop, Ethan Hall, Victor Carbune, Abhinav Rastogi, and Sushant Prakash.
\newblock Rlaif vs. rlhf: Scaling reinforcement learning from human feedback with ai feedback, 2023.
\newblock arXiv:2309.00267.

\bibitem{cheng-aif-2024}
Jing-Cheng Pang, Pengyuan Wang, Kaiyuan Li, Xiong-Hui Chen, Jiacheng Xu, Zongzhang Zhang, and Yang Yu.
\newblock Language model self-improvement by reinforcement learning contemplation, 2023.
\newblock arXiv:2305.14483.

\bibitem{Bai-constitutional-2022}
Yuntao Bai, Saurav Kadavath, Sandipan Kundu, Amanda Askell, Jackson Kernion, Andy Jones, Anna Chen, Anna Goldie, Azalia Mirhoseini, Cameron McKinnon, Carol Chen, Catherine Olsson, Christopher Olah, Danny Hernandez, Dawn Drain, Deep Ganguli, Dustin Li, Eli Tran-Johnson, Ethan Perez, Jamie Kerr, Jared Mueller, Jeffrey Ladish, Joshua Landau, Kamal Ndousse, Kamile Lukosuite, Liane Lovitt, Michael Sellitto, Nelson Elhage, Nicholas Schiefer, Noemi Mercado, Nova DasSarma, Robert Lasenby, Robin Larson, Sam Ringer, Scott Johnston, Shauna Kravec, Sheer~El Showk, Stanislav Fort, Tamera Lanham, Timothy Telleen-Lawton, Tom Conerly, Tom Henighan, Tristan Hume, Samuel~R. Bowman, Zac Hatfield-Dodds, Ben Mann, Dario Amodei, Nicholas Joseph, Sam McCandlish, Tom Brown, and Jared Kaplan.
\newblock Constitutional ai: Harmlessness from ai feedback, 2022.
\newblock arXiv:2212.08073.

\bibitem{minae-reward-2023}
Minae Kwon, Sang~Michael Xie, Kalesha Bullard, and Dorsa Sadigh.
\newblock Reward design with language models.
\newblock In {\em The Eleventh International Conference on Learning Representations}, 2023.

\bibitem{singh-scaling-2023}
Avi Singh, John~D Co-Reyes, Rishabh Agarwal, Ankesh Anand, Piyush Patil, Xavier Garcia, Peter~J Liu, James Harrison, Jaehoon Lee, Kelvin Xu, Aaron~T Parisi, Abhishek Kumar, Alexander~A Alemi, Alex Rizkowsky, Azade Nova, Ben Adlam, Bernd Bohnet, Gamaleldin~Fathy Elsayed, Hanie Sedghi, Igor Mordatch, Isabelle Simpson, Izzeddin Gur, Jasper Snoek, Jeffrey Pennington, Jiri Hron, Kathleen Kenealy, Kevin Swersky, Kshiteej Mahajan, Laura~A Culp, Lechao Xiao, Maxwell Bileschi, Noah Constant, Roman Novak, Rosanne Liu, Tris Warkentin, Yamini Bansal, Ethan Dyer, Behnam Neyshabur, Jascha Sohl-Dickstein, and Noah Fiedel.
\newblock Beyond human data: Scaling self-training for problem-solving with language models.
\newblock {\em Transactions on Machine Learning Research}, 2024.
\newblock Expert Certification.

\bibitem{pmlr-v202-hu23e}
Hengyuan Hu and Dorsa Sadigh.
\newblock Language instructed reinforcement learning for human-{AI} coordination.
\newblock In Andreas Krause, Emma Brunskill, Kyunghyun Cho, Barbara Engelhardt, Sivan Sabato, and Jonathan Scarlett, editors, {\em Proceedings of the 40th International Conference on Machine Learning}, volume 202 of {\em Proceedings of Machine Learning Research}, pages 13584--13598. PMLR, 23--29 Jul 2023.

\bibitem{mehri-eskenazi-2020-unsupervised}
Shikib Mehri and Maxine Eskenazi.
\newblock Unsupervised evaluation of interactive dialog with {D}ialo{GPT}.
\newblock In Olivier Pietquin, Smaranda Muresan, Vivian Chen, Casey Kennington, David Vandyke, Nina Dethlefs, Koji Inoue, Erik Ekstedt, and Stefan Ultes, editors, {\em Proceedings of the 21th Annual Meeting of the Special Interest Group on Discourse and Dialogue}, pages 225--235, 1st virtual meeting, July 2020. Association for Computational Linguistics.

\bibitem{ferron-etal-2023-meep}
Amila Ferron, Amber Shore, Ekata Mitra, and Ameeta Agrawal.
\newblock {MEEP}: Is this engaging? prompting large language models for dialogue evaluation in multilingual settings.
\newblock In Houda Bouamor, Juan Pino, and Kalika Bali, editors, {\em Findings of the Association for Computational Linguistics: EMNLP 2023}, pages 2078--2100, Singapore, December 2023. Association for Computational Linguistics.

\bibitem{wang-etal-2024-incharacter}
Xintao Wang, Yunze Xiao, Jen-tse Huang, Siyu Yuan, Rui Xu, Haoran Guo, Quan Tu, Yaying Fei, Ziang Leng, Wei Wang, Jiangjie Chen, Cheng Li, and Yanghua Xiao.
\newblock {I}n{C}haracter: Evaluating personality fidelity in role-playing agents through psychological interviews.
\newblock In Lun-Wei Ku, Andre Martins, and Vivek Srikumar, editors, {\em Proceedings of the 62nd Annual Meeting of the Association for Computational Linguistics (Volume 1: Long Papers)}, pages 1840--1873, Bangkok, Thailand, August 2024. Association for Computational Linguistics.

\bibitem{lin-chen-2023-llm}
Yen-Ting Lin and Yun-Nung Chen.
\newblock {LLM}-eval: Unified multi-dimensional automatic evaluation for open-domain conversations with large language models.
\newblock In Yun-Nung Chen and Abhinav Rastogi, editors, {\em Proceedings of the 5th Workshop on NLP for Conversational AI (NLP4ConvAI 2023)}, pages 47--58, Toronto, Canada, July 2023. Association for Computational Linguistics.

\bibitem{liu-etal-2023-g}
Yang Liu, Dan Iter, Yichong Xu, Shuohang Wang, Ruochen Xu, and Chenguang Zhu.
\newblock {G}-eval: {NLG} evaluation using gpt-4 with better human alignment.
\newblock In Houda Bouamor, Juan Pino, and Kalika Bali, editors, {\em Proceedings of the 2023 Conference on Empirical Methods in Natural Language Processing}, pages 2511--2522, Singapore, December 2023. Association for Computational Linguistics.

\bibitem{zheng-judge-2023}
Lianmin Zheng, Wei-Lin Chiang, Ying Sheng, Siyuan Zhuang, Zhanghao Wu, Yonghao Zhuang, Zi~Lin, Zhuohan Li, Dacheng Li, Eric~P. Xing, Hao Zhang, Joseph~E. Gonzalez, and Ion Stoica.
\newblock Judging llm-as-a-judge with mt-bench and chatbot arena, 2023.
\newblock arXiv:2306.05685.

\bibitem{sugiyama-jtrans-2022}
Hiroaki Sugiyama, Masahiro Mizukami, Tsunehiro Arimoto, Hiromi Narimatsu, Yuya Chiba, Hideharu Nakajima, and Toyomi Meguro.
\newblock Empirical analysis of training strategies of transformer-based japanese chit-chat systems.
\newblock In {\em 2022 IEEE Spoken Language Technology Workshop (SLT)}, pages 685--691, 2023.

\bibitem{schulman-ppo-2017}
John Schulman, Filip Wolski, Prafulla Dhariwal, Alec Radford, and Oleg Klimov.
\newblock Proximal policy optimization algorithms, 2017.
\newblock arXiv:1707.06347.

\bibitem{rafailov-dpo-2024}
Rafael Rafailov, Archit Sharma, Eric Mitchell, Stefano Ermon, Christopher~D. Manning, and Chelsea Finn.
\newblock Direct preference optimization: Your language model is secretly a reward model, 2023.
\newblock arXiv:2305.18290.

\bibitem{rashkin-etal-2019-towards}
Hannah Rashkin, Eric~Michael Smith, Margaret Li, and Y-Lan Boureau.
\newblock Towards empathetic open-domain conversation models: A new benchmark and dataset.
\newblock In Anna Korhonen, David Traum, and Llu{\'\i}s M{\`a}rquez, editors, {\em Proceedings of the 57th Annual Meeting of the Association for Computational Linguistics}, pages 5370--5381, Florence, Italy, July 2019. Association for Computational Linguistics.

\end{thebibliography}

\end{document}